\documentclass[runningheads]{llncs}
\usepackage[utf8]{inputenc}

\usepackage{eccv}

\usepackage{eccvabbrv}

\usepackage{graphicx}
\usepackage{booktabs}

\usepackage[accsupp]{axessibility}  

\usepackage{hyperref}

\usepackage{orcidlink}

\usepackage[table]{xcolor}

\begin{document}

\title{6Bit-Diffusion: Inference-Time Mixed-Precision Quantization for Video Diffusion Models} 


\titlerunning{~}

\author{Rundong Su\inst{1,2}$^{, *, \ddagger}$,
Jintao Zhang\inst{2}$^{, *}$,
Zhihang Yuan\inst{2}, 
Haojie Duanmu\inst{3}, \\
Jianfei Chen\inst{2}, 
Jun Zhu\inst{2}}


\authorrunning{~}

\institute{\textsuperscript{1}Fudan University \quad \textsuperscript{2}Tsinghua University \quad \textsuperscript{3}Shanghai Jiao Tong University\\
\email{surd.scholar@gmail.com, dcszj@tsinghua.edu.cn}}

\maketitle

\begingroup
\renewcommand{\thefootnote}{*}
\footnotetext{Equal contribution.}
\renewcommand{\thefootnote}{\ddag}
\footnotetext{This work was done during an internship at the TSAIL Group, Tsinghua University.}
\endgroup

\begin{abstract}
  Diffusion transformers have demonstrated remarkable capabilities in generating videos. However, their practical deployment is severely constrained by high memory usage and computational cost. Post-Training Quantization provides a practical way to reduce memory usage and boost computation speed. 
  Existing quantization methods typically apply a static bit-width allocation, overlooking the quantization difficulty of activations across diffusion timesteps, leading to a suboptimal trade-off between efficiency and quality.
  In this paper, we propose a inference time NVFP4/INT8 Mixed-Precision Quantization framework.
  We find a strong linear correlation between a block's input-output difference and the quantization sensitivity of its internal linear layers.
  Based on this insight, we design a lightweight predictor that dynamically allocates NVFP4 to temporally stable layers to maximize memory compression, while selectively preserving INT8 for volatile layers to ensure robustness. 
  This adaptive precision strategy enables aggressive quantization without compromising generation quality.
  Beside this, we observe that the residual between the input and output of a Transformer block exhibits high temporal consistency across timesteps. 
  Leveraging this temporal redundancy, we introduce Temporal Delta Cache (TDC) to skip computations for these invariant blocks, further reducing the computational cost. Extensive experiments demonstrate that our method achieves 1.92$\times$ end-to-end acceleration and $3.32\times$ memory reduction, setting a new baseline for efficient inference in Video DiTs.
  \keywords{Video DiTs \and Mixed-precision quantization \and Efficient Inference}
\end{abstract}

\section{Introduction}

\label{sec:intro}
\noindent Diffusion Transformers (DiTs)~\cite{peebles2023scalable} have revolutionized video generation, achieving remarkable fidelity and temporal consistency~\cite{blattmann2023stable,liu2024sora,ma2024latte}. However, this performance comes with heavy memory and computational costs. For instance, large models like HunyuanVideo will directly cause Out-Of-Memory on consumer devices due to massive parameters~\cite{zhang2025turbodiffusion}. Furthermore, even for relatively smaller models like CogVideoX~\cite{yang2024cogvideox} with two billion parameters generating a 49-frame 1080p video still takes about 22 minutes on an NVIDIA RTX-5090. Such heavy overheads severely limit the fast generation and practical deployment of video DiTs. 

\begin{figure}[!t]
    \centering
    \includegraphics[width=0.99\linewidth]{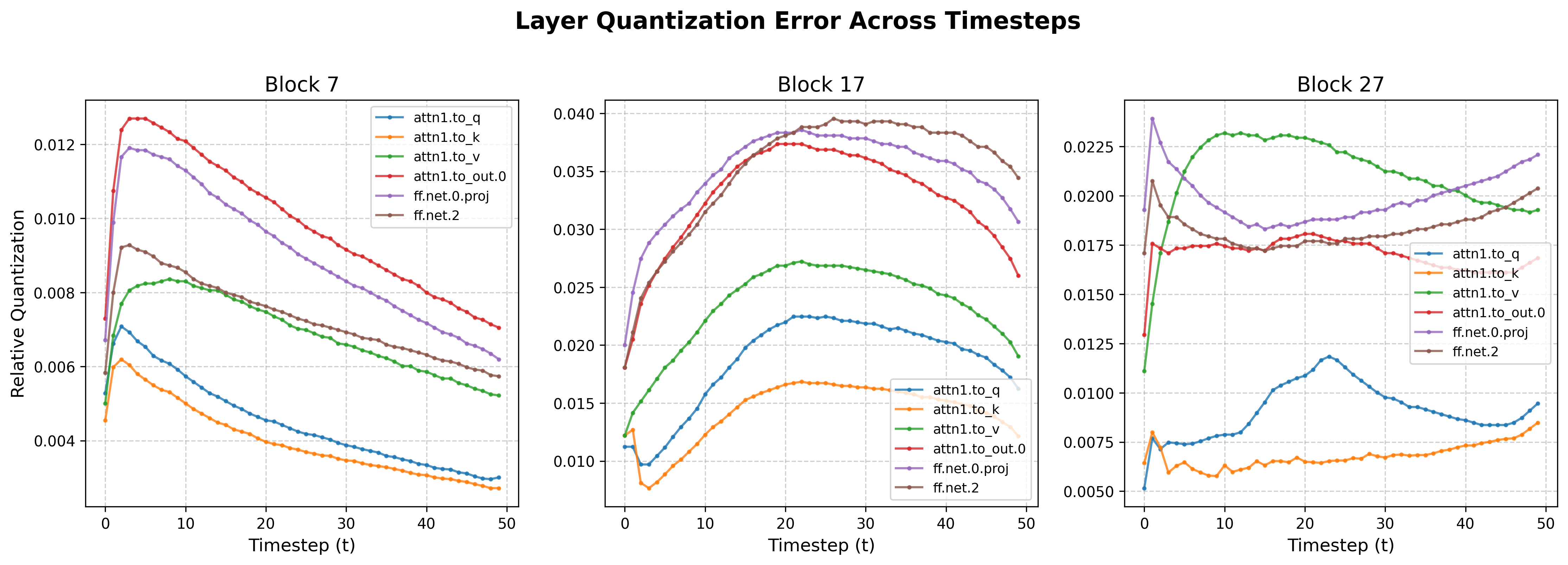}
    \caption{Relative L2 quantization error(defined in Eq.~\ref{eq:r_l2}) of individual linear layers in CogVideoX~\cite{yang2024cogvideox} across denoising timesteps. The severe temporal fluctuations demonstrate that activation sensitivity to quantization is highly dynamic, highlighting the limitations of static mixed-precision policies.}
    \label{fig:quant_error}
\end{figure}

\noindent Model quantization~\cite{jacob2018quantization} serves as a practical method to reduce the memory and computational costs by compressing the weight and activation into low bit-width formats. In particular, Post-training quantization (PTQ) offers a training-free and deployment-friendly solution~\cite{li2023q,he2023ptqd}, but existing methods face significant limits. Uniform quantization~\cite{li2024svdquant,chen2025q} applies a single bit-width to all layers, causing precision loss or insufficient compression. Static mixed-precision methods~\cite{zhao2024vidit,wu2025quantcache} assign layer-specific precision offline and keep these settings fixed during inference. We observe that activation sensitivity to quantization changes drastically across denoising timesteps(as Fig~\ref{fig:quant_error} shows). A static policy either causes severe temporal flickering in sensitive steps or wastes compression opportunities in stable steps.

\noindent To address this, we analyze the denoising process of video DiTs and gain two key insights regarding dynamic quantization sensitivity and temporal redundancy. First, we find a strong linear correlation between a block's input-output difference at the previous timestep and the quantization error of its internal linear layers at the current timestep. If a block shows a large relative input-output difference, its internal layers are highly sensitive to quantization and require higher precision (e.g., INT8). Conversely, layers within stable blocks can safely use ultra-low bit-widths (e.g., NVFP4). This simple linear relationship allows us to predict dynamic precision requirements online with minimal overhead. Second, the residual between the input and output of a Transformer block are highly similar across adjacent timesteps. This temporal consistency indicates heavy computational redundancy, allowing us to skip redundant block computations without degrading video quality.


\noindent Driven by these insights, we propose a unified, training-free acceleration framework that systematically exploits these dynamic characteristics. Specifically, we integrate Dynamic Mixed-Precision Quantization (DMPQ) with a Temporal Delta Cache (TDC). DMPQ dynamically allocates NVFP4 or INT8 precision to activations based on the feature differences from the previous timestep. Complementarily, TDC reuses cached delta updates when temporal similarity is high, skipping redundant block computations. Together, they achieve extreme inference acceleration while maintaining high visual fidelity. Experimental results show that our framework achieves a 1.92× speedup and a 3.32× GPU memory reduction on CogVideoX~\cite{yang2024cogvideox}, while preserving comparable video quality to the full-precision baseline. 
\noindent Our main contributions are summarized as follows:
\begin{itemize}
\item We propose DMPQ, the first dynamic mixed-precision quantization framework utilizing NVFP4 and INT8 for modern GPU architectures (e.g., NVIDIA Blackwell). It allocates precision online based on temporal sensitivity, achieving highly efficient, lossless generation.
\item We introduce TDC, a complementary caching mechanism that exploits temporal redundancy. By selectively skipping expensive block updates, TDC seamlessly synergizes with DMPQ to form a unified spatiotemporal acceleration framework for video DiTs.
\item Extensive experiments on state-of-the-art Video DiTs (e.g., CogVideoX) demonstrate the superiority of our framework. Our method achieves significant inference speedup and memory reduction while maintaining comparable visual quality and temporal consistency to the uncompressed baseline.
\end{itemize}

\section{Related Works}

\subsection{Video Diffusion Transformers (DiTs)}
Diffusion Transformers (DiTs)~\cite{peebles2023scalable} have replaced U-Net~\cite{ronneberger2015u} becoming the standard backbone for visual generation.
Unlike U-Net, DiTs use self-attention~\cite{vaswani2017attention} to better capture long-range dependencies and complex structural relationships.
This shift has led to powerful open-source video models. For example, Open-Sora~\cite{zheng2024open, peng2025open} uses a Spatial-Temporal DiT with a 3D autoencoder for high-quality video generation. CogVideoX~\cite{yang2024cogvideox} uses a 3D full attention mechanism to align text and video well. HunyuanVideo~\cite{hunyuanvideo2025} uses a hybrid architecture with smart attention designs to improve efficiency and video quality.
However, despite the good video quality, video DiTs still facing a lot challenges. The huge number of parameters in these models necessitate huge memory allocation, which significantly strains hardware resources. Beside this, the large matrix multiplications, quadratic self-attention, and iterative denoising steps cause high computational costs. These challenges led to a lot of research in community to improve DiT' efficiency, such as model quantization~\cite{zhao2024vidit, chen2025q, li2024svdquant,wu2025quantcache}, efficient attention~\cite{zhang2025sageattention,zhang2024sageattention2,zhang2025sageattention2++,zhang2025sageattention3,zhang2026sagebwd,zhang2025spargeattention,zhang2026spargeattention2,zhang2025sla,zhang2026sla2,zhangefficient,zhang2025ditfastattnv2}, caching~\cite{chen2024delta, ma2024deepcache, liu2025timestep} and efficient sampling~\cite{song2020denoising,zhang2025turbodiffusion,lu2022dpm,lu2025dpm,lipman2022flow}

\subsection{Model Quantization}


Post-Training Quantization (PTQ) effectively reduces memory and compute costs for DiTs. By compressing full-precision weights and activations into low-bit formats, it reduce the memory footprint and accelerates computation on modern GPUs without retraining.

\noindent Early methods primarily focused on Large Language Models (LLMs) or U-Nets. For example, SmoothQuant~\cite{xiao2023smoothquant} addressed the channel outlier in activations by transfer quantization difficulty from activations to weights using channel-wise scaling. QuaRot~\cite{ashkboos2024quarot} smoothing outlizers by apply randomized Hadamard rotations to weights and activations. Recently, researchers adapted these to DiTs. Q-Diffusion~\cite{li2023q} and PTQ4DiT~\cite{wu2024ptq4dit} design calibration methods for the denoising process. Q-DiT~\cite{chen2025q} uses fine-grained group quantization. ViDiT-Q~\cite{zhao2024vidit} uses a mixed-precision strategy, giving different bit-widths to different layers using metric-decoupled  sensitivity.

\noindent Although these methods improve quantization quality, they have two main limits. First, most methods~\cite{zhao2024vidit,wu2025quantcache} use static precision policies. They fix the bit-widths across all timesteps. However, we observe that a model's sensitivity changes over time. Static policies cause flickering in sensitive stages or waste compression chances in stable stages. Second, many low-bit formats (like INT4) lack hardware support on the newest GPUs. For example, the NVIDIA Blackwell architecture removes INT4 Tensor Core support and introduces FP4. Currently, no video DiT quantization methods have integrated this new hardware-native format.

\subsection{Diffusion Caching}
Feature caching accelerates DiTs by exploiting the high temporal redundancy of feature maps and attention states across adjacent timesteps. It reuses these cached features to skip redundant computations without retraining.

\noindent Early methods like DeepCache~\cite{ma2024deepcache} and FORA~\cite{selvaraju2024fora} use fixed schedules to skip redundant blocks. Later methods improve this by dynamically deciding when to skip computations based on step-to-step changes. For instance, AdaCache~\cite{kahatapitiya2025adaptive} skips redundant steps by checking feature differences. TeaCache~\cite{liu2025timestep} estimates output differences using timestep embeddings to determine when to reuse the cache. EasyCache~\cite{wang2014easycache} monitors runtime stability to make the same cache-reuse decision. $\Delta$-DiT~\cite{chen2024delta} caches feature residuals (deltas) instead of direct features to prevent information loss.

\noindent While effective, these caching methods typically operate in isolation from model quantization. They treat caching and quantization as orthogonal things, missing the strong link between temporal stability and quantization sensitivity. Furthermore, simply combining them causes accumulated quantization errors (drift), which ruins video quality. Therefore, we need a unified framework that considers both caching and model quantization together.

\section{Preliminaries of Quantization}
Quantization~\cite{nagel2021white} compresses weights and activations into low bit-width formats to reduce memory footprints and computational overhead. Given a full-precision tensor $X$, the general quantization and dequantization processes are formulated as:
\begin{equation}
        \text{Quant}: X_q = \text{clip}\left(\left\lfloor\frac{X}{s}\right\rceil + z, q_{\text{min}}, q_{\text{max}}\right) \quad  
        \text{Dequant}: \hat{X} = (X_q - z) \cdot s
\end{equation}

where $X_q$ and $\hat{X}$ denote the quantized and dequantized tensors, respectively. Here $s$ is quantization scaling factor, $z$ is the zero-point, and $[q_{\text{min}}, q_{\text{max}}]$ defines the representable range of the target bit-width. The specific calculation of $s$ and $z$ depends on the chosen quantization scheme:\\
\textbf{Integer Quantization.} For unsymmetrical INT8 quantization, the tensor is mapped to an unsigned range $[0, 255]$ to accommodate skewed distributions. The scaling factor and zero-point are calculated as $s = (X_{\max} - X_{\min}) / 255$ and $z = -\lfloor X_{\min} / s \rceil$. When the tensor distribution is roughly symmetric around zero (e.g., neural network weights), symmetrical INT8 quantization is employed. It drops the zero-point ($z=0$) and maps values to $[-128, 127]$ with $s = \max(|X|) / 127$.\\
\textbf{NVFP4 Quantization.} Recent hardware architectures introduce micro-scaling block-level formats like NVFP4 for extreme compression. NVFP4 consists of 1 sign bit, 2 exponent bits, and 1 mantissa bit (E2M1), with a maximum representable value of 6.0. It projects continuous values into a discrete FP4 set using a shared FP8 scaling factor $s = \max(|X|) / 6.0$ for a contiguous block (e.g., 16 elements):
\begin{equation}
    \text{Quant}: X_q = \text{CastToFP4}_{\text{E2M1}}\left(\frac{X}{s}\right)  \quad
    \text{Dequant}:  \hat{X} = X_q \cdot s
\end{equation} 
\noindent where \textit{$\text{CastToFP4}$} maps normalized values to the nearest representable FP4 magnitude.

\section{Methods}
\begin{figure}[!t]
    \centering
    \includegraphics[width=0.99\linewidth]{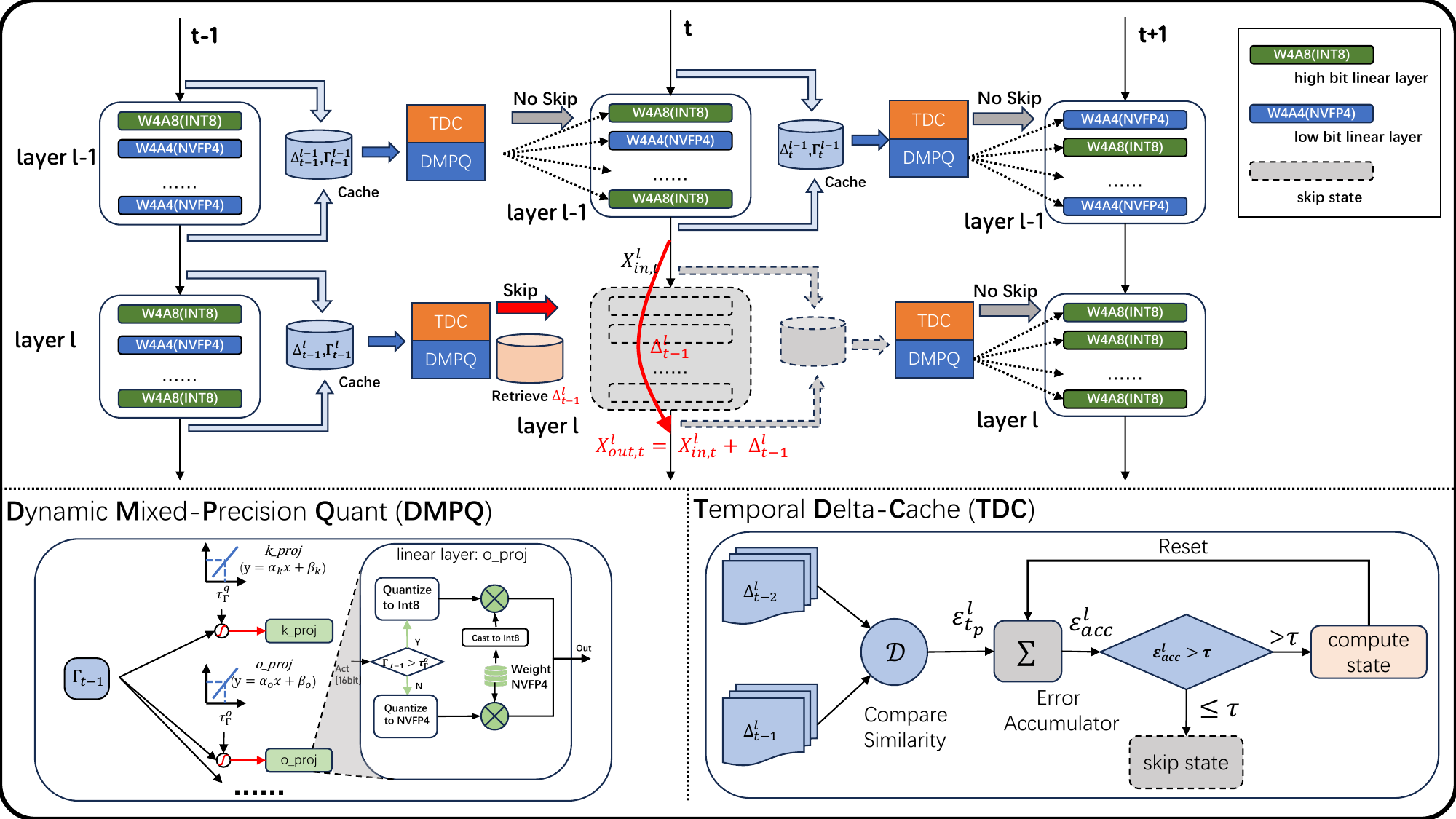}
    \caption{Overview of our proposed methods. We using DMPQ to decide the  quantization bits of each linear layer activation based on the block preceding timestep input output relative L1 loss $\Gamma$. And using TDC to decide whether to skip block computation based on preceding two timesteps block updates $\Delta$.}
    \label{fig:placeholder}
\end{figure}

\subsection{Dynamic Mixed-Precision Quantization (DMPQ)}
\label{sec:dynamic_mpq}

\begin{figure}[htbp]
    \centering
    \includegraphics[width=0.99\linewidth]{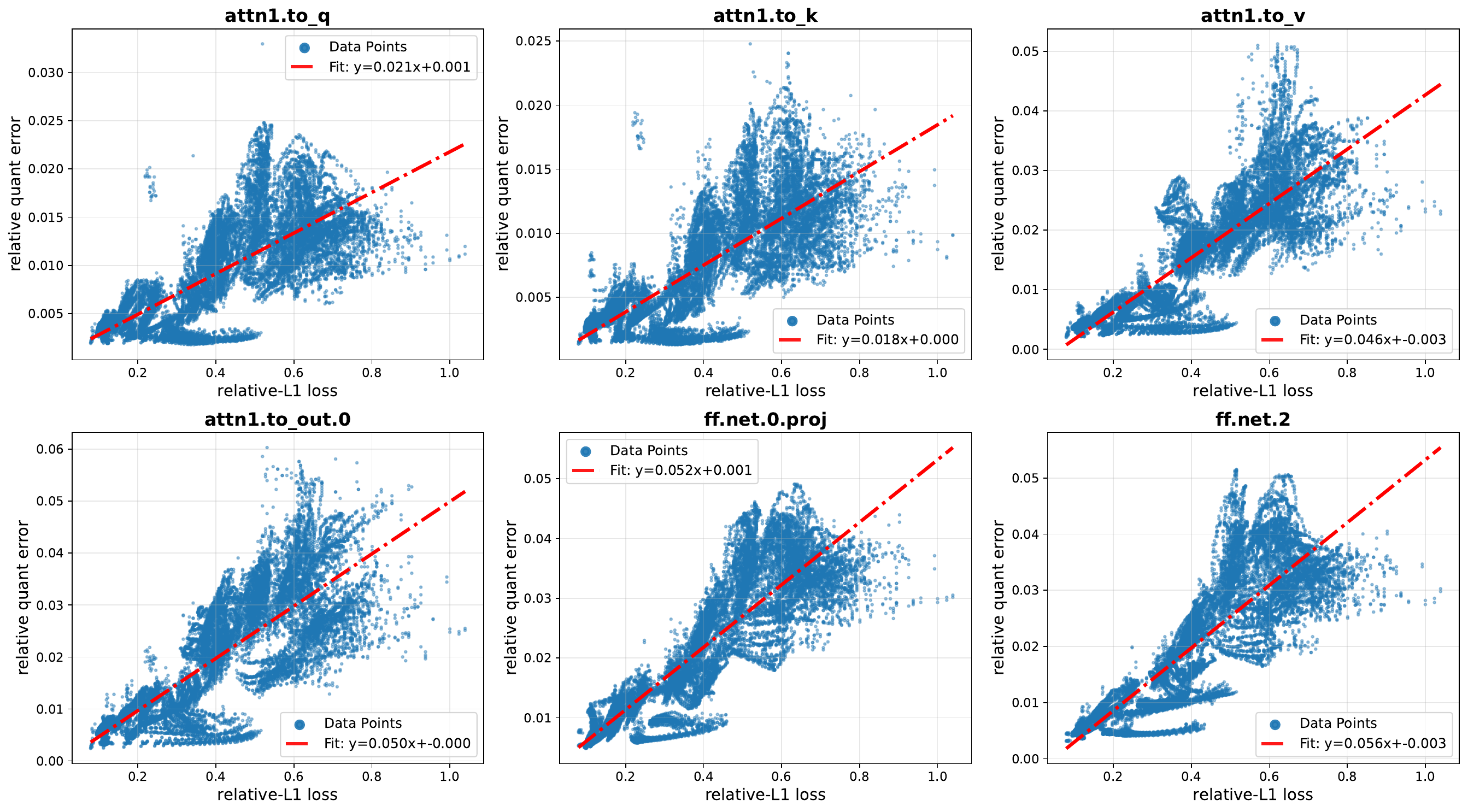}
    \caption{The linear relationship between the block-level input-output relative L1 distance at the previous timestep ($\Gamma_{t-1}$) and the layer-wise relative quantization error ($E_{rel}$) at the current timestep. The data points are collected by executing the diffusion model over a calibration set. For each block, we extract the time-shifted valid data pairs across all timesteps to perform linear regression ($y = \alpha x + \beta$) for each internal linear layer.}
    \label{fig:DMPQ_linear}
\end{figure}

\textbf{Motivation and Observation.}
While full NVFP4 activation quantization suffers from severe quality degradation due to outliers, uniformly employing INT8 underutilizes the acceleration capabilities of modern GPUs. Although a mixed-precision approach balances 8-bit and 4-bit assignments, existing static mix precision methods are suboptimal for Video DiTs due to the severe fluctuation of individual layer sensitivities across timesteps (Fig.~\ref{fig:quant_error}).
We address this by leveraging a novel observation: a linear layer's quantization sensitivity exhibits a strong linear relationship with the  previous timestep relative difference between the input and output of its located block.
Specifically, within a transformer block, the relative quantization error of a linear layer at timestep $t$ is linearly correlated with the relative L1 distance between the block's output and input at timestep $t-1$.
As illustrated in Fig.~\ref{fig:DMPQ_linear}, the relative quantization errors across different linear layers can be roughly modeled as a simple linear function of the previous timestep's block-level transformation magnitude. \\

\noindent \textbf{Formulation of Metrics.} To formalize this, we use the relative L1 distance between the block input $X_{t-1}$ and output $Y_{t-1}$ at timestep $t-1$ to define the block-level transformation magnitude, termed as $\Gamma_{t-1}$:
\begin{equation}
    \Gamma_{t-1} = \frac{||Y_{t-1} - X_{t-1}||_1}{||X_{t-1}||_1}
\end{equation}
To decouple the quantization sensitivity from the varying activation magnitudes across timesteps and layers, we adopt a scale-invariant metric. Specifically, the relative quantization error $E_{rel}$ for a specific linear layer inside this block is measured by the normalized L2 distance between the block full-precision output $O$ and  quantized output $O_q$(where only this specific layer is quantized):
\begin{equation}
    \label{eq:r_l2}
    E_{rel} = \frac{||O - O_q||_2}{||O||_2}
\end{equation}

\noindent \textbf{Layer-wise Linear Predictive Modeling.} 
Building on this observation, for any specific linear layer (e.g., attention Q/K/V/O proj or FFN layers) within a block, its relative quantization error $E_{rel}$ can be modeled as a linear function of the block's input-output relative difference $\Gamma_{t-1}$:
\begin{equation}
    E_{rel} = \alpha \cdot \Gamma_{t-1} + \beta
\end{equation}
where the layer-specific slope $\alpha$ and intercept $\beta$ are pre-fitted offline using a small calibration set. Computing $\Gamma_{t-1}$ once per block to determine the precision routing for all its internal layers introduces negligible runtime overhead. \\

\noindent \textbf{Dynamic Mixed-Precision Routing.} During inference, we predefine an acceptable relative error threshold $\tau_{rel}$. Using our linear model, we derive a layer-specific relative L1 distance threshold $\tau_{\Gamma}$ for each projection by inverting the equation:
\begin{equation}
    \tau_{\Gamma} = \frac{\tau_{rel} - \beta}{\alpha}
\end{equation}
At timestep $t$, we compare the block's computed $\Gamma_{t-1}$ against $\tau_{\Gamma}$ for each of its internal linear layers to assign quantization bits to the activation. This routing mechanism is formulated as Eq.~\ref{eq:routing}:
\begin{equation}
\label{eq:routing}
    A_{bits} =
\begin{cases}
    \text{INT8} & \text{if } \Gamma_{t-1} > \tau_{\Gamma} \\
    \text{NVFP4}  & \text{if } \Gamma_{t-1} \le \tau_{\Gamma}
\end{cases}
\end{equation}
To strictly minimize the memory footprint, all weights are quantized to NVFP4 offline. During the forward pass, if a layer's $A_{bits}$ is routed to INT8, its corresponding weights are cast to INT8 on-the-fly solely to satisfy GEMM data type requirements. \\


\noindent \textbf{Outlier Smoothing.} To mitigate severe activation outliers, we utilize an online Block Hadamard Transform. Traditional global Hadamard transformations are often disrupted by non-linear operations (e.g., GELU~\cite{hendrycks2016gaussian}) due to their reliance on offline weight fusion. To avoid this limitation, we apply a Fast Hadamard Transform (FHT)~\cite{fino1976unified, tseng2024quip} over local activation  blocks ($B=128$). This localized design restricts the rotation complexity to $\mathcal{O}(\log B)$ per element and allows the operation to be seamlessly fused into our custom quantization kernels. Consequently, activation outliers are effectively redistributed on-the-fly with negligible overhead, after which the smoothed activations are quantized using either the NVFP4 format or per-block symmetric INT8.


\subsection{Temporal Delta Cache}
\label{sec:TDC}

\begin{figure}[htbp]
    \centering
    \includegraphics[width=0.99\linewidth]{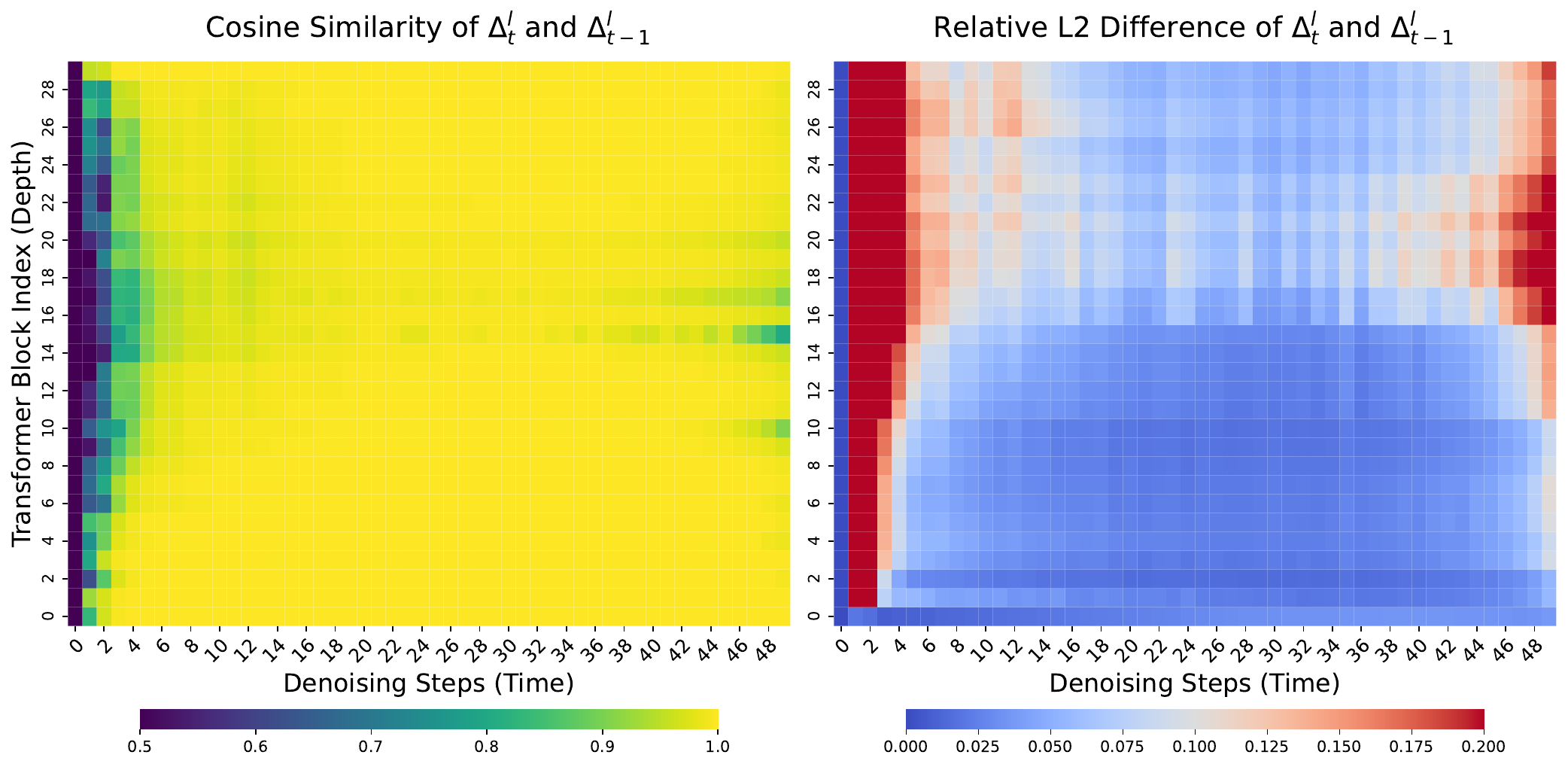}
    \caption{Temporal redundancy of Transformer block updates in Video DiTs. We visualize the Cosine Similarity (left) and Relative L2 Difference (right) of the residual updates ($\Delta_t^l$ and $\Delta_{t-1}^l$) between adjacent timesteps. The inherently high similarity across most timesteps and layers motivates our adaptive Temporal Delta Cache.}
    \label{fig:tdc_motivation}
\end{figure}

\textbf{Empirical Observation.} In Video DiTs, the residual deltas of transformer blocks exhibit significant similarity across adjacent timesteps. Let $X_{in, t}^l$ denote the unified input (concatenated text and visual hidden states) of the $l$-th block at timestep $t$. The block's forward pass can be abstracted as:
\begin{equation}
    X_{out, t}^l = X_{in, t}^l + \Delta_t^l,
\end{equation}
where $\Delta_t^l$ represents the residual delta at timestep $t$. As illustrated in Fig.~\ref{fig:tdc_motivation}, we visualize the Cosine Similarity and Relative L2 Difference between adjacent deltas ($\Delta_t^l$ and $\Delta_{t-1}^l$). The results demonstrate strong temporal consistency throughout the majority of the diffusion process, where the deltas remain highly correlated ($\Delta_t^l \approx \Delta_{t-1}^l$). Furthermore, we observe that this temporal redundancy inherently varies across different transformer layers. This consistent yet layer-dependent temporal similarity directly motivates our adaptive caching mechanism to skip redundant block computations.

\noindent \textbf{Theoretical Insight.} Diffusion sampling corresponds to solving a Probability Flow ODE (PF-ODE)~\cite{song2020score}. As denoising progresses, the ODE trajectory curvature decreases, leading to a smoother velocity field~\cite{karras2022elucidating, lu2022dpm}. Consequently, the network outputs become locally linear across adjacent timesteps. This inherent smoothness physically explains our observation that $\Delta_t^l \approx \Delta_{t-1}^l$, and justifies using the historical discrepancy $\mathcal{D}(\Delta_{t-1}^l, \Delta_{t-2}^l)$ to estimate the current prediction error $\mathcal{E}_t^l$.

\noindent \textbf{Predictive caching mechanism.} Building upon this observation, when a block's update trajectory exhibits high stability, we can reuse the historical update to significantly reduce the computational overhead. However, since the diffusion inference is strictly online, the current update $\Delta_t^l$ is inaccessible prior to computation. Therefore, we utilize the similarity between the previous two updates to predict the stability of the current step. We define the prediction error $\mathcal{E}_t^l$ using a generalized distance function $\mathcal{D}$ to measure the discrepancy between the historical updates:
\begin{equation}
    \label{eq:tdc_func}
    \mathcal{E}_t^l = \mathcal{D}(\Delta_{t-1}^l, \Delta_{t-2}^l)
\end{equation}
In our default implementation, we use $1 - \text{CosSim}(\Delta_{t-1}^{l}, \Delta_{t-2}^{l})$ as $\mathcal{D}$, though other metrics such as relative-L2 distance can also be applied depending on the speed-quality trade-off. \\
\textbf{Error-Guided Cache Switching.} While adjacent timesteps updates are highly consistent, continuous caching inevitably introduces approximation drift. To control this theoretical error and safely refresh the cache, we propose an Error-Guided Cache Switching mechanism governed by an accumulated error metric $\mathcal{E}_{acc}^l$.Let $t_p$ denote the most recent timestep at which the block was fully computed, and $\mathcal{E}_{t_p}^l$ be the exact prediction error calculated at that step. At the end of any timestep $t-1$, we dynamically update the accumulated error to evaluate the cache viability for the next timestep $t$:
\begin{equation}
    \label{eq:TDC_err_accum}
    \mathcal{E}_{acc}^l \leftarrow \begin{cases} \mathcal{E}_{t_p}^l & \text{if } \mathcal{S}_{t-1}^l = \text{Compute} \\ \mathcal{E}_{acc}^l + \mathcal{E}_{t_p}^l + \rho & \text{if } \mathcal{S}_{t-1}^l = \text{Skip} \end{cases}
\end{equation}
where $\rho$ is a constant penalty factor. This formulation guarantees that the penalty $\rho$ is strictly excluded when evaluating the initial skip ($t = t_p + 1$), and is only introduced to penalize unobserved drift during continuous caching.Based on this metric, the execution state $\mathcal{S}_t^l$ for the current timestep $t$ is strictly determined by:
\begin{equation}
\label{eq:TDC_state_judge}
\mathcal{S}_{t}^l = \begin{cases} \text{Skip} & \mathcal{E}_{acc}^l \leq \tau \text{ and } t - t_p \le N_{max} \\ \text{Compute} & \text{otherwise} \end{cases}
\end{equation}
where $\tau$ is the global cache threshold, and $t - t_p$ represents the consecutive skip count.When $\mathcal{S}_t^l = \text{Skip}$, the block computation is skipped, and the output is approximated by reusing the cached delta: $X_{out, t}^l \approx X_{in, t}^l + \Delta_{t_p}^l$. Caching these residual deltas introduces only a small memory overhead, as they can be quantized to ultra-low precision formats such as NVFP4. Conversely, when $\mathcal{S}_t^l = \text{Compute}$, the system executes the full Transformer block. This action automatically refreshes the cache and resets the accumulated error in the subsequent step, effectively clearing the approximation drift. 

\subsection{Purified Cache Refresh}
\label{sec:synergy}

\noindent \textbf{Quality Degradation in Naive DMPQ-TDC Combination.} While DMPQ and TDC independently yield substantial acceleration, naively combining them into Video DiTs causes severe video quality degradation. This degradation fundamentally comes from the temporal accumulation of quantization noise. Regardless of the dynamically assigned bit-width (NVFP4 or INT8), DMPQ inherently introduces a single-step quantization error $\epsilon_q$ into the computed delta.

\noindent \textbf{Formulation of Error Accumulation.} To formalize this, let the actual computed residual with quantization noise be $\tilde{\Delta}_{t_p}^l = \Delta_{t_p}^l + \epsilon_q$. As introduced in Sec.~\ref{sec:TDC}, if a block is skipped for $N$ consecutive steps, the approximated output becomes:
\begin{equation}
    \label{eq:quant_cache_drift}
    X_{out, t+N}^l \approx X_{in, t+N}^l + N \cdot \tilde{\Delta}_{t_p}^l = X_{in, t+N}^l + N \cdot \Delta_{t_p}^l + N \cdot \epsilon_q
\end{equation}
Eq.~\ref{eq:quant_cache_drift} reveals that the single-step error $\epsilon_q$ is linearly amplified by skip count $N$. To prevent this accumulation, it is critical to minimize quantization noise in the computed $\Delta_t$ before it is cached.


\noindent \textbf{Outlier-Aware Cache Purification.} To fully mitigate the temporal accumulation of quantization noise, the $\Delta_t$ written into the cache must be as pure as possible. Since extreme outliers easily corrupt the cache, we first evaluate quantization difficulty by spatially sampling the input to estimate its outlier ratio $R_{outlier}(X) = \frac{\max(|X|)}{\mathrm{mean}(|X|)}$. If $R_{outlier}$ exceeds a threshold $\tau_{outlier}$, the layer skips quantization and uses full precision (FP16/BF16), ensuring the cache is refreshed with purified, high-fidelity features. Conversely, if the activation is quantization-friendly, we allocate lower precision formats via DMPQ~(Sec.~\ref{sec:dynamic_mpq}). However, when a block resumes computation after a skip ($\mathcal{S}_{t-1}^l = \text{Skip}$), the metric $\Gamma_{t-1}$ is missing, which disables the dynamic routing in Eq.~\ref{eq:routing}. As a safe fallback for this specific timestep, we simply assign INT8 precision to all linear layers in the block. This mechanism effectively isolates temporal recurrence from quantization errors.


\section{Experiments}
\subsection{Experiments Setup}



\noindent \textbf{Models and Datasets.} We evaluate our framework on two scale variants of CogVideoX~\cite{yang2024cogvideox}. To model the relationship between layer-wise quantization sensitivity and the preceding block's variation magnitude, we use a calibration set of 100 randomly sampled prompts from EvalCrafter~\cite{liu2024evalcrafter}.

\noindent \textbf{Baselines.} We compare our framework against widely adopted quantization techniques, including SmoothQuant~\cite{xiao2023smoothquant} and QuaRot~\cite{ashkboos2024quarot}, as well as the current state-of-the-art post-training quantization (PTQ) method for video generation, ViDiT-Q~\cite{zhao2024vidit}.

\noindent \textbf{Evaluation Metrics.} To comprehensively assess generation quality, visual fidelity, and spatial-temporal consistency, we utilize two standardized benchmarking frameworks: VBench~\cite{huang2024vbench} and EvalCrafter~\cite{liu2024evalcrafter}. For the VBench suite, following prior works~\cite{zhao2024vidit}, we select 8 major dimensions to ensure a thorough assessment. From the EvalCrafter pipeline, we report CLIPSIM and CLIP-Temp to measure text-video alignment and temporal semantic consistency, respectively. Furthermore, we employ DOVER~\cite{wu2023exploring} to evaluate video quality from both aesthetic and technical perspectives, and compute the Flow-score to assess fine-grained temporal consistency.

\noindent \textbf{Implementation Details.} For all evaluations, we follow the official CogVideoX configurations to ensure a fair comparison. Videos are generated using the DDIM scheduler~\cite{song2020denoising} with 50 denoising timesteps and a classifier-free guidance (CFG)~\cite{ho2022classifier} scale of 6.0. All experiments are conducted on a single NVIDIA RTX-5090 GPU.For our proposed framework, we set the dynamic routing threshold $\tau_{\Gamma}=0.015$, the TDC penalty factor $\rho=0.001$, the maximum consecutive cache steps $N_{max}=2$, and the global cache threshold $\tau=0.003$. To strictly prevent quantization noise accumulation during generation, the outlier threshold $\tau_{outlier}$ in Sec.~\ref{sec:synergy} is set to $25$.

\subsection{Effectiveness Results}
\begin{figure}[!t]
    \centering
    \includegraphics[width=0.99\linewidth]{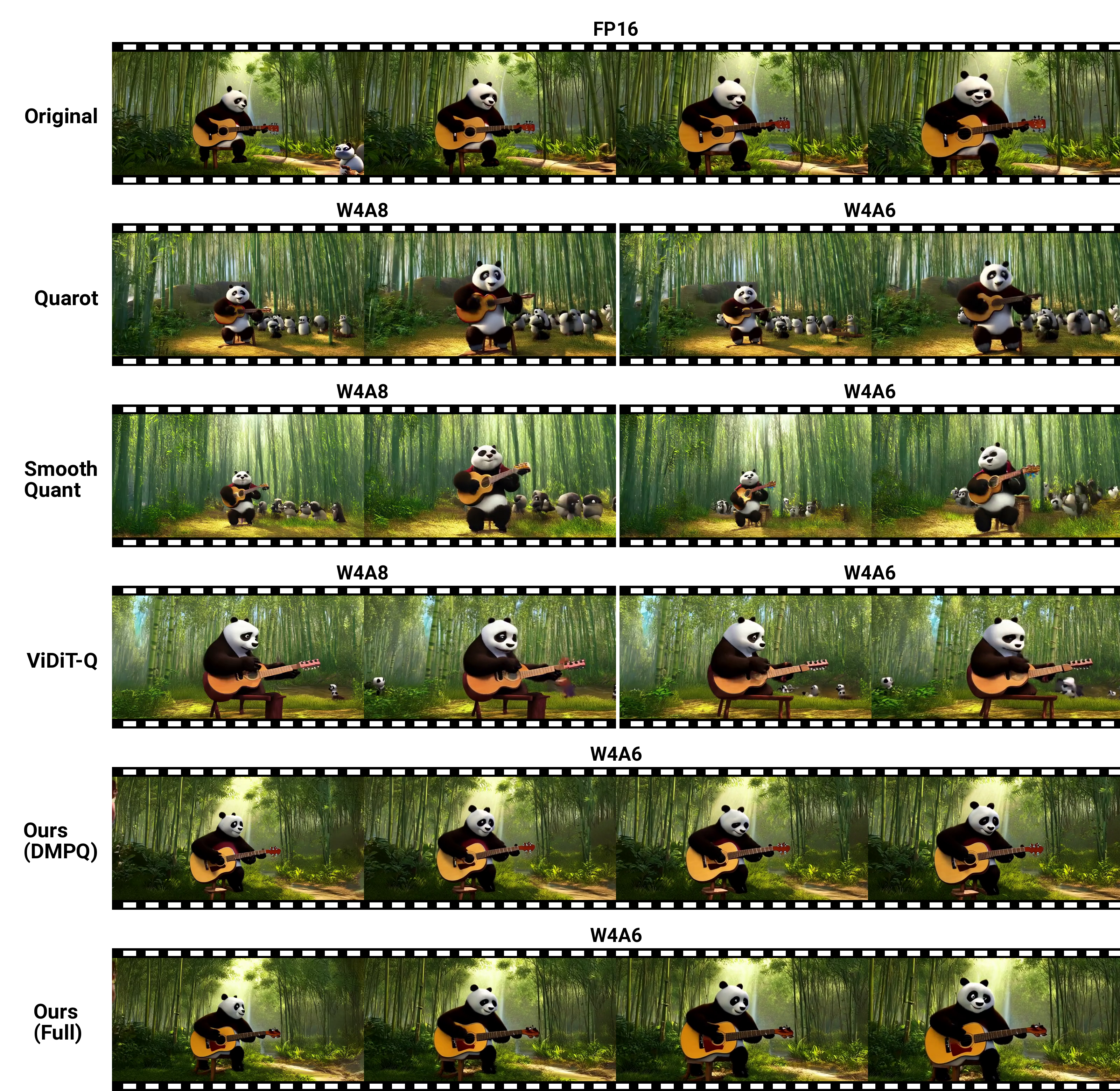}
    \caption{Visual comparisons between ours and FP16 baseline~\cite{yang2024cogvideox}), together with quantization methods~\cite{ashkboos2024quarot}, SmoothQuant~\cite{xiao2023smoothquant} and ViDiT-Q~\cite{zhao2024vidit}}
    \label{fig:qualitive_compare}
\end{figure}

\textbf{Quantitative Evaluation}: We evaluate our framework on the VBench benchmark~\cite{huang2024vbench} using both CogVideoX-2B and the larger CogVideoX-5B models. As shown in Table~\ref{tab:vbench} under the aggressive W4A6 setting, traditional static quantization methods suffer from severe performance degradation, especially on the larger 5B model (e.g., ViDiT-Q drops to an Aesthetic Quality of 0.4433). In contrast, our Dynamic Mixed-Precision Quantization (DMPQ) demonstrates remarkable robustness.
For the 2B model, DMPQ achieves an Aesthetic Quality of 0.5437, significantly outperforming all W4A6 baselines and even matching or exceeding state-of-the-art static W4A8 methods like SmoothQuant (0.5332) and ViDiT-Q (0.5332). By leveraging the linear correlation between temporal instability and quantization sensitivity, DMPQ adaptively protects volatile layers with INT8 while aggressively compressing stable layers to NVFP4. This dynamic allocation allows us to maintain a superior Overall Consistency of 0.2474 on the 2B model and 0.2492 on the 5B model, closing the gap with the full-precision baselines.
\begin{table*}[!t]
\centering
\caption{Performance comparison of various quantization methods on VBench using CogVideo-2B and CogVideo-5B.}
\label{tab:vbench}
\resizebox{\linewidth}{!}{
\begin{tabular}{lccccccccc}
\toprule[1.5pt]
Method & Bit-width & Aesthetic & BG. & Overall & Dynamic & Subject & Imaging & Scene & Motion \\
&(W/A) & Quality & Consist. & Consist. & Degree & Consist. & Quality & Consist. & Smooth. \\
\midrule[1.5pt]
\multicolumn{10}{c}{\textbf{CogVideo-2B}} \\
\midrule[1.5pt]
Original & W16/A16 & 0.5464 & 0.9500 & 0.2423 & 0.7639 & 0.9491 & 0.5713 & 0.3590 & 0.9714 \\
\midrule
SmoothQuant & W4A8 & 0.5332 & 0.9441 & 0.2473 & 0.6806 & 0.9319 & 0.5460 & 0.3343 & 0.9680 \\
QuaRot      & W4A8 & 0.5158 & 0.9427 & 0.2383 & 0.5139 & 0.9202 & 0.5258 & 0.2682 & 0.9715 \\
ViDiT-Q     & W4A8 & 0.5332 & 0.9449 & 0.2464 & 0.6528 & 0.9272 & 0.5534 & 0.3372 & 0.9692 \\
SmoothQuant & W4A6 & 0.5010 & 0.9444 & 0.2348 & 0.4444 & 0.9141 & 0.5118 & 0.2391 & 0.9714 \\
QuaRot      & W4A6 & 0.5300 & 0.9408 & 0.2455 & 0.6805 & 0.9277 & 0.5406 & 0.3481 & 0.9667 \\
ViDiT-Q     & W4A6 & 0.4745 & 0.9470 & 0.2202 & 0.3750 & 0.9086 & 0.4701 & 0.1766 & 0.9725 \\
\rowcolor{gray!15} Ours (DMPQ) & W4A6 & 0.5437 & 0.9455 & 0.2474 & 0.6806 & 0.9282 & 0.5511 & 0.3307 & 0.9691 \\
\midrule[1.5pt]
\multicolumn{10}{c}{\textbf{CogVideo-5B}} \\
\midrule[1.5pt]
Original & W16/A16 & 0.5922 & 0.9570 & 0.2595 & 0.6944 & 0.9505 & 0.6079 & 0.4608 & 0.9760 \\
\midrule
SmoothQuant & W4A8 & 0.5148 & 0.9502 & 0.2355 & 0.4722 & 0.9230 & 0.5115 & 0.2522 & 0.9796 \\
QuaRot      & W4A8 & 0.5371 & 0.9496 & 0.2448 & 0.5278 & 0.9347 & 0.5428 & 0.3685 & 0.9779 \\
ViDiT-Q     & W4A8 & 0.4525 & 0.9606 & 0.2036 & 0.2222 & 0.9111 & 0.4355 & 0.0879 & 0.9818 \\
SmoothQuant & W4A6 & 0.4950 & 0.9499 & 0.2351 & 0.4167 & 0.9162 & 0.5009 & 0.1933 & 0.9791 \\
QuaRot      & W4A6 & 0.5345 & 0.9485 & 0.2435 & 0.4861 & 0.9304 & 0.5377 & 0.3517 & 0.9775 \\
ViDiT-Q     & W4A6 & 0.4433 & 0.9617 & 0.2034 & 0.1667 & 0.9110 & 0.4260 & 0.0784 & 0.9819 \\
\rowcolor{gray!15} Ours (DMPQ) & W4A6 & 0.5724 & 0.9598 & 0.2492 & 0.6389 & 0.9454 & 0.5637 & 0.4513 & 0.9705 \\
\bottomrule[1.5pt]
\end{tabular}
}
\end{table*}
To further maximize acceleration without degrading visual fidelity, we combine DMPQ with TDC using our proposed method. The evaluation results are shown in Table~\ref{tab:vbench2}. While standalone DMPQ achieves a $1.36\times$ speedup, combining it with TDC achieves a $1.92\times$ speedup with almost no degradation in generation quality.
\begin{table*}[!t]
\centering
\caption{Performance evaluation of integrating TDC into DMPQ on VBench~\cite{huang2024vbench}.}
\label{tab:vbench2}
\resizebox{\linewidth}{!}{
\begin{tabular}{lcccccccccc}
\toprule
Method & Bit-width & Speedup & Aesthetic & BG. & Overall & Dynamic & Subject & Imaging & Scene & Motion \\
&(W/A) & & Quality & Consist. & Consist. & Degree & Consist. & Quality & Consist. & Smooth. \\
\midrule
Original & W16A16 & $1.00\times$ & 0.5464 & 0.9500 & 0.2423 & 0.7639 & 0.9491 & 0.5713 & 0.3590 & 0.9714 \\
\midrule
Ours(DMPQ)        & W4A6   & $1.36\times$ & 0.5437 & 0.9455 & 0.2474 & 0.6806 & 0.9282 & 0.5511 & 0.3307 & 0.9691 \\
Ours(Full)    & W4A6$^*$ & $1.92\times$ & 0.5366 & 0.9421 & 0.2492 & 0.7083 & 0.9255 & 0.5369 & 0.3307 & 0.9691 \\
\bottomrule
\end{tabular}
}
{\raggedright \scriptsize $^*$ \textbf{Note:} With TDC, activations in skipped blocks are treated as 0-bit width, resulting in an overall average activation bit-width of 6.035 ($\approx$ 6 bits).\par}
\end{table*}

\noindent \textbf{Qualitative Comparison.} Fig~\ref{fig:qualitive_compare} compares our method against state-of-the-art PTQ baselines (Quarot~\cite{ashkboos2024quarot}, SmoothQuant~\cite{xiao2023smoothquant}, and ViDiT-Q~\cite{zhao2024vidit}) at W4A8 and W4A6. Existing methods suffer from severe visual degradation. Specifically, Quarot and SmoothQuant exhibit semantic drift and hallucination (e.g., generating non-existent pandas), with SmoothQuant further losing spatiotemporal consistency at W4A6. While ViDiT-Q avoids gross hallucinations, it introduces significant local geometric distortions, such as the structural deformation of the guitar. In contrast, our DMPQ method, and DMPQ interged with TDC and PDR, effectively mitigates these quantization errors. It preserves both fine-grained details and global spatiotemporal consistency, achieving visual fidelity comparable to the original FP16 model even under W4A6.

\subsection{Memory and Latency Analysis}
Our proposed framework significantly reduces the model memory footprint by $3.32\times$ compared to the BF16/FP16 baseline while accelerating inference. To eliminate the computational overhead caused by online activation quantization and the block Hadamard transform, we develop custom hardware CUDA kernels to fuse these operations directly into the preceding layers. This makes the quantization overhead practically negligible, achieving a $1.36\times$ end-to-end speedup. Note that this acceleration is naturally bounded by the attention mechanism, which consumes over half of the total execution time. Furthermore, integrating our Temporal Delta Cache (TDC) to skip redundant block computations further boosts the overall speedup to $1.92\times$ with minimal degradation in visual quality.

\subsection{Ablation Study}
We conduct an ablation study on the EvalCrafter~\cite{liu2024evalcrafter} benchmark to evaluate the individual contributions of Dynamic Mixed Precision Quantization (DMPQ), Temporal Delta Cache (TDC), and Purified Delta Refresh (PDR). Table~\ref{tab:ablation} summarizes the results across visual (\textit{VQA-A/T}), semantic (\textit{CLIPSIM/TEMP}), and temporal (\textit{FLOW SCORE}) metrics. The results are summarized in Table~\ref{tab:ablation}.

\noindent \textbf{Impact of Dynamic Mixed Precision Quantization (DMPQ).}
As shown in Table~\ref{tab:ablation}, naive uniform W4A4 quantization leads to severe degradation across multiple dimensions, notably causing the Flow Score to drop from 5.8389 to 4.2518. Uniform W4A8 mitigates this but yields suboptimal memory compression. Ablation 1 introduces our DMPQ strategy. By dynamically routing layers to NVFP4 or INT8 based on their temporal instability ($\frac{\| \text{Output} - \text{Input} \|_1}{\| \text{Input} \|_1}$), DMPQ effectively protects volatile layers. This mechanism achieves a VQA-A of 75.5332, successfully outperforming the FP16 baseline in static visual fidelity, thereby demonstrating the necessity of instability-aware precision allocation.

\begin{table}[htbp]
    \centering
    \caption{\textbf{Ablation Study of Proposed Components.} Evaluation of Dynamic Mixed Precision Quantization (DMPQ), Temporal Delta Cache (TDC), and Purified Delta Refresh (PDR). }
    \label{tab:ablation}
    \resizebox{\linewidth}{!}{
    \begin{tabular}{l | ccc | ccccc}
        \toprule
        \textbf{Method} & \textbf{DMPQ} & \textbf{TDC} & \textbf{PDR} & \textbf{VQA-A} $\uparrow$ & \textbf{VQA-T} $\uparrow$ & \textbf{CLIP} $\uparrow$ & \textbf{CLIP-T} $\uparrow$ & \textbf{Flow Score} $\uparrow$ \\
        \midrule
        Original (FP16) & - & - & - & 74.6175 & 77.7548 & 0.1824 & 0.9973 & 5.8389 \\
        Uniform W4A4 & - & - & - & 68.9105 & 73.6116 & 0.1818 & 0.9974 & 4.2518 \\
        Uniform W4A8 & - & - & - & 71.2093 & 78.6070 & 0.1859 & 0.9972 & 5.3140 \\
        Ablation 1 & \checkmark & & & 75.5332 & 77.8404 & 0.1814 & 0.9976 & 5.3266 \\
        Ablation 2 &  & \checkmark & & 72.2671 & 78.1517 & 0.1838 & 0.9975 & 5.7602 \\
        Ablation 3 & \checkmark & \checkmark &  & 73.7243 & 77.2833 & 0.1835 & 0.9978 & 4.7271 \\
        Ablation 4 & \checkmark & \checkmark & \checkmark & 76.1173 & 76.6990 & 0.1848 & 0.9973 & 5.5417 \\
        \bottomrule
    \end{tabular}
    }
\end{table}


\noindent \textbf{Integration of Temporal Delta Cache and Purified Delta Refresh.}
Ablation 2 evaluates TDC in an unquantized (FP16) setting. While maintaining high temporal consistency (Flow Score 5.7602) by skipping redundant computations, it offers no memory relief. Ablation 3 naively combines TDC with Dynamic Mixed Precision Quantization (DMPQ). This induces severe error drift, as quantization noise compounds with cached deltas across frames, plummeting the Flow Score to 4.7271. Ablation 4 resolves this critical issue via the Purified Delta Refresh (PDR). By selectively routing hard-to-quantize or temporally uncertain layers to higher precision, PDR effectively breaks the error accumulation chain. This targeted fallback preserves the memory benefits of quantization, restores the Flow Score to 5.5417, and achieves the highest visual quality (VQA-A 76.1173), demonstrating an optimal efficiency-fidelity trade-off.

\section{Conclusion}
In this paper, we propose 6Bit-Diffusion to reduce the high memory and computational costs of Video Diffusion Transformers. Our framework consists of three main parts. First, we introduce Dynamic Mixed-Precision Quantization (DMPQ), which dynamically allocates NVFP4 to stable layers and INT8 to sensitive layers based on their temporal instability. Second, we design a Temporal Delta Cache (TDC) to skip redundant block computations across adjacent time steps. Third, to prevent quantization errors from accumulating in the cache, we propose a Purified Delta Refresh (PDR) mechanism. Extensive experiments on CogVideoX show that our method is highly effective, accelerating inference by 1.92x and reducing memory usage by 3.32x, while maintaining the original video quality.

\clearpage  


%
%
\bibliographystyle{splncs04}
\bibliography{main}
\end{document}